\documentclass[runningheads]{llncs}
\usepackage{graphicx}

\usepackage{tikz}
\usepackage{comment}
\usepackage{amsmath,amssymb} % define this before the line numbering.
\usepackage{color}
\usepackage{multirow}

\usepackage{graphicx}
\usepackage{amsmath}
\usepackage{amssymb}
\usepackage{booktabs}
\usepackage{amsmath}
\usepackage{algorithm}
\usepackage{algpseudocode} 
\usepackage[font=small,labelfont=bf]{caption}
\usepackage{multicol} 
\usepackage[rightcaption]{sidecap}
\usepackage{subfig}

\usepackage[accsupp]{axessibility}  % Improves PDF readability for those with disabilities.

\begin{document}
\pagestyle{headings}
\mainmatter
\def\ECCVSubNumber{7543}  % Insert your submission number here

\title{Generating Natural Images with Direct Patch Distributions Matching}

\titlerunning{Generating Natural images with Direct Patch Distributions Matching}
% If the paper title is too long for the running head, you can set
% an abbreviated paper title here

\author{Ariel Elnekave \and Yair Weiss}%
\authorrunning{A. Elnekave \and Y. Weiss.}

\institute{The Hebrew University of Jerusalem \\
\email{\{Ariel.Elnekave, Yair.Weiss\}@mail.huji.ac.il}}

%******************
\maketitle
\begin{figure}[H]
    \makebox[\textwidth][c]{\includegraphics[width=1\textwidth]{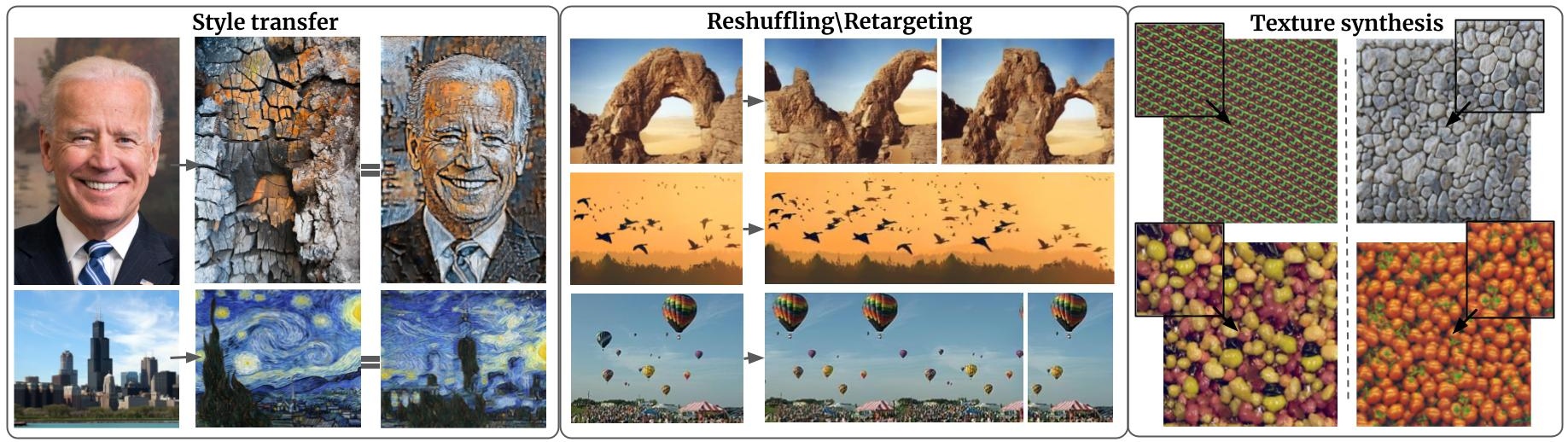}}%
    \caption{By efficiently matching the distribution of patches between images we can solve a broad spectrum of single-image generative tasks without training a per-image GAN or computing patch nearest neighbors.}
   \label{fig:Teaser}
\end{figure}
\begin{abstract}
Many traditional computer vision algorithms generate realistic images by requiring that each patch in the generated image be similar to a patch in a training image and vice versa. Recently, this classical approach has been replaced by adversarial training with a patch discriminator. The adversarial approach avoids the computational burden of finding nearest neighbors of patches but often requires very long training times and may fail to match the distribution of patches.

In this paper we leverage the Sliced Wasserstein Distance to develop an algorithm that explicitly and efficiently minimizes the distance between patch distributions in two images.
Our method is conceptually simple, requires no training and can be implemented in a few lines of codes. On a number of image generation tasks we show that our results are often superior to single-image-GANs,  and can generate high quality images in a few seconds.
Our implementation is publicly available at~{\color{magenta}{https://github.com/ariel415el/GPDM}}.
\end{abstract}

\section{Introduction}

In a wide range of computer vision problems (e.g. image retargeting, super-resolution, novel view synthesis) an algorithm needs to generate a realistic image as an output. A classical approach to ensuring that the output image appears realistic is based on {\em local patches}~\cite{BIDIRECTIONAL,PATCH_MATCH,QUILTING}: if each patch in the output image is similar to a patch in a training image, we can assume that the generated image will be realistic. Similarly, if each patch in the generated image is similar to a patch in a Van-Gogh painting, we can assume that the generated image captures the ``style'' of Van-Gogh. This insight led to a large number of papers over the past two decades that use patch nearest neighbors to ensure high quality image outputs~\cite{QUILTING,CONTEXTUAL,ImageAnalogies,BLIND_DEBLURING}.

One observation shared by successful methods based on patches is that the similarity between patches should be {\em bidirectional}: it is not enough to require that each patch in the generated image be similar to a patch in the training image. Consider a generated image that consists of many repetitions of a single patch from a Van-Gogh image: even though each patch in the generated image is similar to a patch in the target image, no one would consider such a generated image to capture the ``style'' of Van-Gogh. In order to rule out such solutions, the bidirectional similarity (BDS) method used in~\cite{BIDIRECTIONAL,PATCH_MATCH,GPNN} also requires that each patch in the training image be similar to a patch in the generated image. As we show in section~\ref{bds-seq}, while bidirectional similarity indeed helps push the distribution of patches in the generated image towards that of the target image, it still falls short of matching the distributions. Furthermore, optimizing the BDS loss function requires finding nearest neighbors of patches and this is both memory and computation intensive (given M patches in each image, BDS is based on an $M^2$ matrix of similarities between all pairs of patches).

In recent years, these classical, patch-based approaches have been overtaken by Generative Adversarial Networks (GANs) and related methods~\cite{iizuka2017globally,SINGAN,INGAN,Wang_2018_CVPR,ConSinGAN,OneShotGAN}. In the adversarial approach, a discriminator is trained to classify patches at different scales as "real" or "fake" and it can be shown that under certain conditions training with an adversarial loss and a patch discriminator is equivalent to minimizing the distance between the distribution of patches in the generated image and the training image~\cite{WGAN,GAN}. The requirement that the generated image has the same {\em distribution} over patches as the target image addresses the limitation of early patch-based approaches that simply required that each patch in the generated image be similar to a patch in the training image. Indeed Single-Image GANs~\cite{INGAN,SINGAN} have yielded impressive results in generating novel images that have approximately the same distribution over patches as the target image.

Despite the considerable success of GAN-based methods, they have some notable disadvantages. While there are theoretical guarantees that globally optimizing the adversarial objective is equivalent to optimizing the distance between distributions, in practice GAN training often suffers from "mode collapse"~\cite{srivastava2017veegan} and the generated image may contain only few types of possible patches. Furthermore, GAN training is computationally intensive and a separate generator needs to be trained for different image tasks. 

In this paper we leverage the previously proposed Sliced Wasserstein Distance~\cite{SWD_1,SWD_1.5,SWD_2} to develop an algorithm that explicitly and efficiently minimizes the distance between patch distributions in two images without the need to compute patch nearest neighbors. 
Our method is conceptually simple, requires no training and can be implemented in a few lines of codes. On a number of image generation tasks we show that our results are often superior to single-image-GANs,  and can generate high quality images in a few seconds.

\section{Distances between distributions} \label{sec:distributions metrics}

Given $M$ patches in two images, how do we compute the distance between the distribution of patches in the two images?
The Wasserstein (or Earth Movers) distance between two distributions $P,Q$ is defined as:

\begin{align}
W(P,Q)= \inf_{\gamma \in \Pi(P,Q)} E_{x,y \sim \gamma} \|x-y\|
\end{align}

where $\Pi(P,Q)$ denotes the set of joint distributions whose marginal probabilities are $P,Q$. Intuitively, $\Pi$ can be thought of as a soft correspondence between samples in $P$ and $Q$ and so the Wasserstein distance is the average distance between corresponding samples with the optimal correspondence. Calculating this optimal correspondence is computationally intensive ($O(M^{2.5})$\cite{SWD_1}) making it unsuitable for use as a loss function that we wish to optimize for many iterations.

The sliced Wasserstein distance (SWD) makes use of the fact that for one dimensional data, the optimal correspondence can be solved by simply sorting the samples and so the distance between two samples of size $M$ can be computed in $O(M \log M)$. For a projection vector $w$ define $P^{w}$ as the distribution of samples from $P$ projected in direction $w$, the Sliced Wasserstein Distance is defined as:

\begin{align}
SWD(P,Q) = E_{w} W(P^w,Q^w)
\end{align}

%It is easy to see that $SWD(P,Q)$ satisfies the metric axioms~\cite{SWD_1}. In particular it is equal to zero if and only if $P=Q$ and is positive otherwise. 

Where the expectation is over random unit norm vectors $\omega$. 

%During optimization we can obtain an unbiased sample of the gradient of $SWD(P,Q)$ by randomly choosing $k$ projection vectors $w_i$ and computing the gradient of:

%\begin{align}
%\tilde{SWD}(P,Q) = \frac{1}{k} \sum_i W(P^{w_i},Q^{w_i})
%\end{align}

\subsection{Properties and Comparisons}
\label{bds-seq}

As mentioned in the introduction, many classical approaches to comparing patch probability distributions are based on bidirectional similarity. Suppose we are given a set of samples $\{p_i\},\{q_j\}$ from two distributions $P,Q$ the Bidirectional Similarity (BDS) is defined as:
\[
BDS(P,Q) = \frac{1}{M} \sum_i \min_j \|p_i-q_j\| + \frac{1}{M} \sum_j \min_i \|q_j - p_i\|
\]
The first term ("coherence") measures the average distance between a patch in $\{p_i\}$ and its closest patch in $\{q_j\}$ and the second term ("completeness") measures the average distance between a patch in $\{q_i\}$ and its closest patch in $\{p_j\}$. Thus two images are judged to be similar if each patch in one image has a close match in the second image and vice versa.

Kolkin et al,~\cite{STROTSS} used a closely related measure which they called the ``Relaxed Earth Movers Distance'' (REMD): 
\[
REMD(P,Q) = \max (\frac{1}{M} \sum_i \min_j \|p_i-q_j\| , \frac{1}{M} \sum_j \min_i \|q_j - p_i\|)
\]
Again, two images are judged to be similar if each patch in one image has a close match in the second image and vice versa.

Thus $SWD(P,Q)$, $BDS(P,Q)$ and $REMD(P,Q)$ are all methods to measure the similarity between the patch distributions. Why should one method be preferred over the others? The following theorem shows that neither BDS nor REMD can be considered as distance metric between distributions, while SWD can.

{\bf Theorem:} $SWD(P,Q)=0$ if and only if $P=Q$. On the other hand for both BDS and REMD, there exist an infinite number of pairs of distributions $P,Q$ that are arbitrarily different (i.e. $W(P,Q)$ is arbitrarily large) and yet $BDS(P,Q)=0$ and $REMD(P,Q)=0$.

{\bf Proof:} The fact that $SWD(P,Q)=0$ if and only if $P=Q$ follows from the fact that the Wasserstein distance is a metric~\cite{SWD_1}. To see that neither BDS nor REMD are metrics, note that any two discrete distributions that have the same support will satisfy $BDS(P,Q)=0$ and $REMD(P,Q)=0$ regardless of the densities on the support. This is because  a single sample in one distribution can serve as an exact match for an arbitrarily large number of samples in the other distribution. For example, suppose $P$ and $Q$ are both distributions over the set $\{0,a\}$ for some constant $a$ and $P(0)=\epsilon, Q(0)= 1 - \epsilon$. For any $\epsilon$ ,$BDS(P,Q)=REMD(P,Q)=0$ (since all samples from $P$ will have an exact matching sample in $Q$ and vice-versa) even though as $\epsilon \rightarrow 0$ $W(P,Q) \rightarrow a$. By increasing $a$ we can increase $W(P,Q)$ arbitrarily. $\blacksquare$

To illustrate the difference between SWD, BDS and REMD in the context of image patches, consider three images of sky and grass, each with 1000 patches. Images A and B, have 999 sky patches and one grass patch, while image C has 1 sky patch and 999 grass patches. Note that when comparing A and B, and when comparing B and C {\em both coherence and completeness losses will be zero.} This means that BDS and REMD will consider A and B (which have the same patch distribution) to be as similar as B and C (which have very different distributions). In contrast, the SWD will be zero if and only if the two distributions are identical, and $SWD(A,B)$ will be much lower than $SWD(B,C)$. 

\subsection{SWD for image patch distributions}
SWD has been previously suggested for use as a training loss for different image processing methods~\cite{SWD_1,SWD_2,SWD_AE,SWD_GAN}. Here we point out that using SWD to measure the similarity of patch distributions in two images allows us to efficiently optimize the patch distribution similarity.

As pointed out in~\cite{SWD_1.5}, while exact SWD requires an integral over all unit norm filters $w$, an {\em approximate} SWD can be obtained by considering a set of $k$ random unit vectors, $\{w_i\}$:
\[
\tilde{SWD}(P,Q) = \frac{1}{k} \sum_i W(P^{w_i},Q^{w_i})
\]
While for any fixed set of $k$ vectors the approximate SWD is not the same as the exact SWD, by taking a gradient of the approximate SWD we obtain an {\em unbiased} estimate of the gradient of the exact SWD. The fact that the estimate is unbiased means that by changing the set of $k$ random vectors at each iterations, we can efficiently optimize the exact SWD, just as is done in training of neural networks with stochastic gradient descent. 

Our additional observation is that for a single projection $w$, the calculation of $\tilde{SWD}$ requires convolving the two images with $w$, sorting the two convolved images with respect to their value and then calculating the L1 distance between the sorted vectors. Similarly, the derivative of the $\tilde{SWD}$ with respect to an image requires a second convolution of the thresholded and sorted difference image with a flipped version of $w$. Combining this observation with the result of~\cite{SWD_1.5} means that a stochastic gradient update with respect to the SWD of the patch distribution in two  images can  be performed with two convolutions and a sorting operation.

Our full algorithm, which we call Generative Patch Distribution Matching (GPDM) is given in Algorithm~\ref{algorithm-1}.  Note that it allows us to optimize the difference between patch distributions in two images {\em without finding patch nearest neighbors} and at a complexity of $O(M \log M)$. 
 
%The unbiased estimate of the distance between the patch distributions in the two images is then simply the L1 distance between the sorted vectors and an unbiased estimate of the gradient of the image with respect to the SWD loss can be obtained by another convolution of the thresholded and sorted difference image with a flipped version of the filter. Thus {\em an SGD update of all pixels in the image with respect to the loss can be computed with two convolutions.}  

\begin{algorithm}[H]
    % \captionof{algorithm}{GPDM Module} 
    \hspace*{\algorithmicindent} \textbf{Input}: Target image x, initial guess $\hat{y}$, \\
    \hspace*{\algorithmicindent} learning rate $\beta$  \\
    \hspace*{\algorithmicindent} \textbf{Output}: Optimized image y
    \begin{algorithmic}[1]
        \State $y\leftarrow\hat{y}$
        \While{not converged}
         \State $L \leftarrow 0$
          \For{i=1,k}
            \State $\omega\sim N(0,\sigma I)$ 
            \State $\omega\leftarrow unflat(\frac{\omega}{\|w\|})$
            \State $p\leftarrow flat(conv2d(x,\omega))$
            \State $q\leftarrow flat(conv2d(y,\omega))$
            \State $L\leftarrow L+ \frac{1}{k} |sort(p) -  sort(q)|$ 
            \EndFor
            \State $y\leftarrow y- \beta\nabla_{y}L$
        \EndWhile
    \end{algorithmic} 
\caption[]{A pseudo-code of the GPDM module where SWD over sets of patches in two images is computed and differentiated through. The "flat/unflat" operators reshapes a tensor into a vector and vice-versa.}
\label{algorithm-1}
\end{algorithm}

%In order to reduce variance, we use a fixed small number of random projections. The full algorithm is described in Alg.~\ref{algorithm-1}.

\section{Method}\label{sec:Method}
Our method uses the same multi scale structure as previous works~\cite{SINGAN,GPNN,ImageAnalogies}. At each level an initial guess is transformed into an output image which is either used as an initial guess for the next level or is the final output. More formally, given a target image $x$ we build an image pyramid out of it $(x_0, x_1,....x_n)$ specified with a downscale ratio $r<1$ and a minimal height for the coarsest level. We start with an initial guess $\hat{y_n}$ of the same size as $x_n$ and at each level $i$ we optimize the initial guess using algorithm~\ref{algorithm-1} to minimize its patch-SWD with $x_i$. The optimization output $y_i$ is a final output or up-scaled by $\frac{1}{r}$ to serve as an initial guess for the next level, $i+1$, optimization. The first initial guess can be a blurred version of the target, a color map or simple pixel noise.

The same learning rate and number of Adam steps is used to optimize SWD at all scales. Images are normalized to [-1, 1] and the optimized image values are clipped to this interval at the end of each level optimization. 

\begin{figure}
    \centering
    \includegraphics[width=\textwidth]{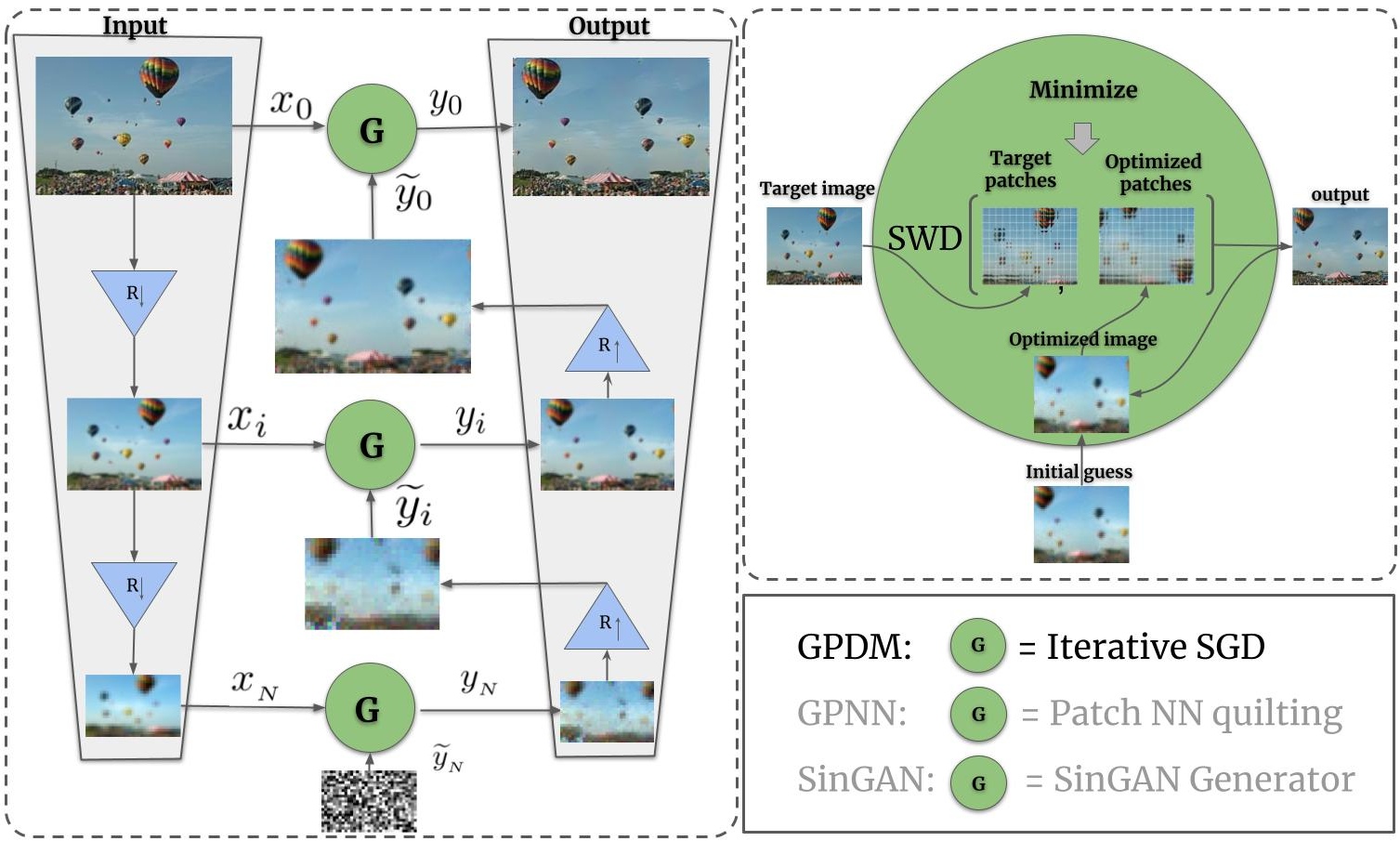}
    \caption{{\bf Left:} GPDM’s multi-scale architecture: At each scale, $i$, an image is optimized to have similar patch distribution as the target $x_i$. {\bf Right:} The generation module $G$  is an optimization process of the differentiable patch-distribution metric $SWD(x,y)$. Triangles marked with \textbf{R} stand for up/downscale.}
    \label{fig:Overview}
\end{figure}

Figure~\ref{fig:Overview} (based on a similar figure in~\cite{GPNN}) shows the overall coarse to fine structure and the minimization process at each scale. 

%%%%%%%%% Experiments
\section{Experiments}

In this section we conduct experiments to compare GPDM to other methods. We first compare the synthesis quality on a number of different single image generation tasks and later compare the algorithms in terms of their run-time efficiency.

The methods we compare to are SinGAN\cite{SINGAN} as well as a recent method, GPNN~\cite{GPNN} which approximately optimizes the bidirectional similarity between the generated image and the target image. GPNN generates a new image by copying patches from the training image in a coarse to fine manner: at each iteration it searches for patches in the training image that are closest to the current estimated image and then aggregates these patches to form a new image. By construction, this method achieves high coherence (since all patches in the new image are copied from the training image) and completeness is encouraged by transforming distances into similarities using a free parameter $\alpha$. When $\alpha \rightarrow \infty$ the patch with maximal similarity is the patch in the second image with minimal L2 distance, but for small $\alpha$ patches in the second image that have already been used will receive low similarity.  

Our method, SINGAN and GPNN all use exactly the same coarse-to-fine strategy but as shown in figure~\ref{fig:Overview}, the difference is in the form of the generator used at each scale.  While SinGAN~\cite{SINGAN} approximately minimizes the KL-divergence between patch distributions and GPNN~\cite{GPNN} approximately minimizes the bidirectional similarity, our method directly minimizes the SWD between the output and target patch distribution.

 The configuration used in each applications of GPDM differ in only the content and size of the first initial guess from which the generation starts. We used a patch size of 7, 300 gradient steps and 64 random projections for all experiments. For more hyper-parameters and details please refer to the supplementary.
 
\subsection{Synthesis quality}

\begin{figure}[hb]
    \centering
    \includegraphics[width=\linewidth]{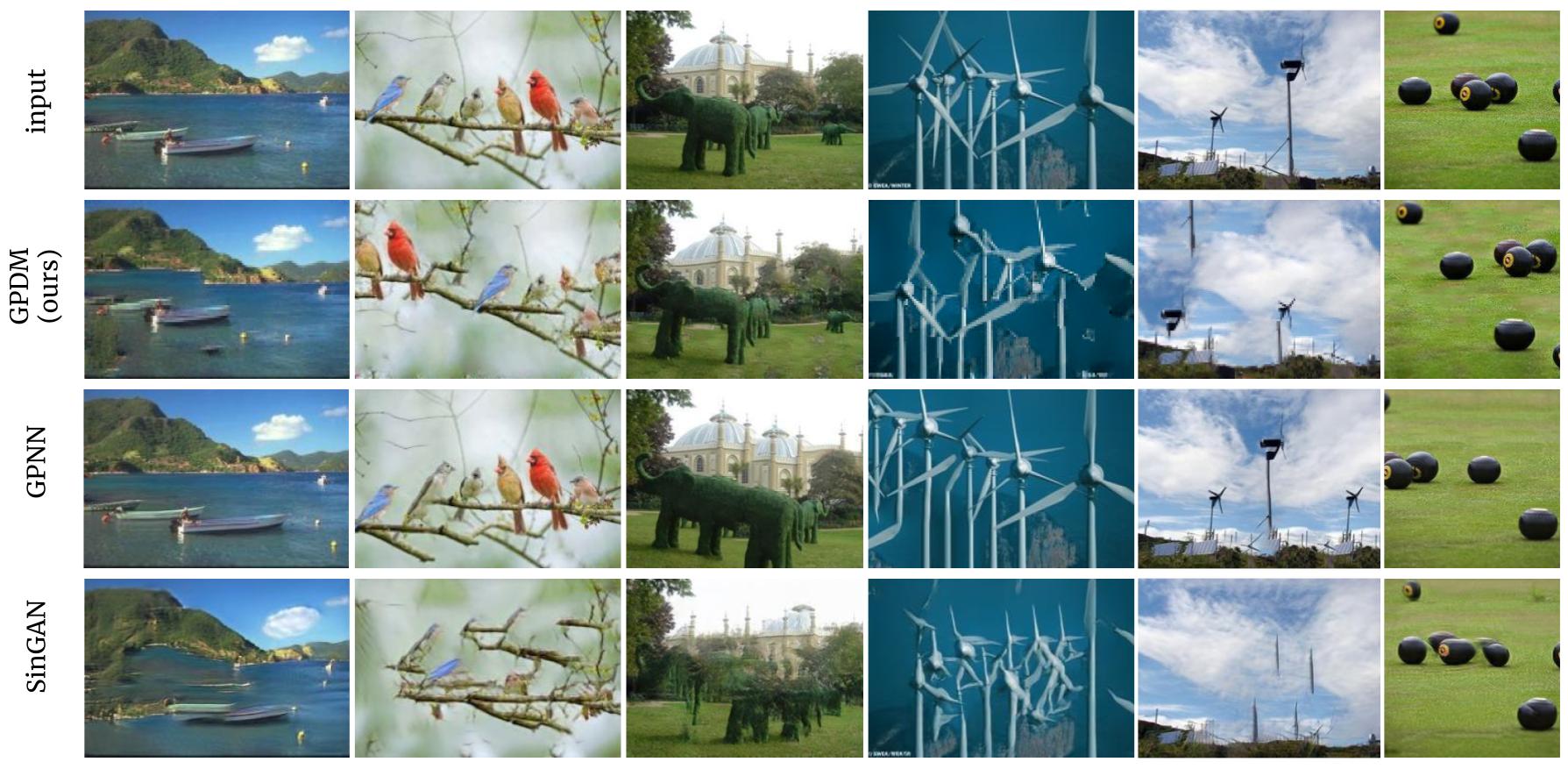}
    \caption{Image reshuffling on images from the Places50 and SIGD16 selected by~\cite{GPNN}.}
\label{fig:Reshuffling}
\end{figure}

\textbf{Image reshuffling:} We first compare the three algorithms on the image reshuffling task task.
Given a natural image the task is to generate more images from the same scene which are different from the original one but show the same scene and are visually coherent. For this task, we start the optimization from a small noise image. Figure~\ref{fig:Reshuffling} visually compares our method to \cite{GPNN,SINGAN} on the same images from Figure 4 in \cite{GPNN}. It can be seen that our method provides comparable visual quality to that of GPNN, and both methods generate more realistic and artifact-free images compared to SinGAN. Some of the images in figure~\ref{fig:Reshuffling} generated by GPDM do show more artifacts compare to GPNN but they are also much less similar to the reference image.

\begin{table}[]
    \centering
    \begin{tabular}{|c|c|c|c|c|c|c|c|}
    %\toprule
    \hline
    Dataset & Method & SFID$\downarrow$ & Diversity$\uparrow$ & Dataset & Method & SFID$\downarrow$  & Diversity$\uparrow$  \\
   % \midrule
   \hline
    \multirow{3}{*}{Places5} & Ours & 0.068 & 0.56 & \multirow{3}{*}{SIGD16} & Ours & 0.069  & 0.67\\
     & GPNN & 0.065 & 0.5  & & GPNN & 0.122 & 0.52 \\
     & SINGAN & 0.082 & 0.5  & & SINGAN & 0.172 & 0.49 \\ 

    %\bottomrule
    \hline
    \end{tabular}
    \caption[]{Qualitative Comparison image-reshuffling.}
    \label{tab:Reshuffling results}
\end{table}

Table~\ref{tab:Reshuffling results} shows a quantitative comparison between our method and \cite{GPNN,SINGAN} using the  SFID  metric (a full reference image quality metric described in \cite{SINGAN})  and also using 
diversity, i.e.  the normalized per-pixel standard deviation over 50 generated images\cite{SINGAN}. Good results should have high diversity (i.e. the generated images are not all identical to the target image) and low SFID (i.e. the generated images are statistically similar to the training image). 

We used the Places50 and SIGD16 datasets from~\cite{GPNN,SINGAN}. We recomputed the SIFID scores for SinGAN and GPNN using the published generated images sets from each paper's supplementary. Our scores differ slightly from those reported in~\cite{GPNN}, presumably due to different implementations. The numerical scores are consistent with the visual inspection: our method is comparable to GPNN in terms of image quality and diversity and both methods outperform SinGAN.

\begin{figure}
 \centering
    \includegraphics[width=0.9\textwidth]{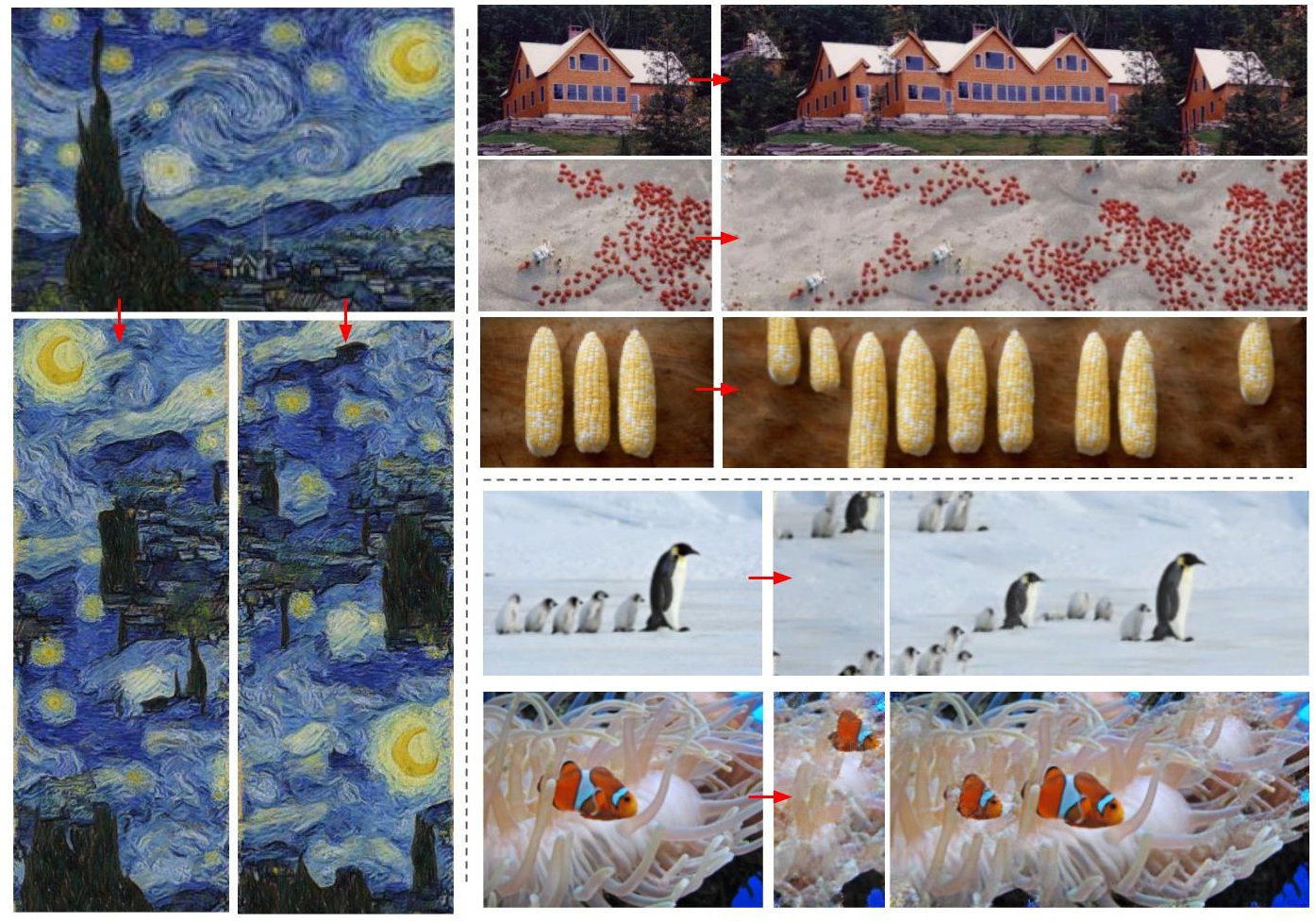}
    \caption{Image retargeting results.}
    \label{fig:Retargeting}
\end{figure}
    
\textbf{Retargeting}: Figure~\ref{fig:Retargeting} shows the results of our method in retargeting images into various aspect ratios. We start the optimization from a stretched version of the target's smallest pyramid level that matches the desired aspect ratio.

In such tasks, where the number of patches in the output and target images differ, we duplicate randomly selected patches from the smaller image so that we compute SWD on two equaly large sets of patches.

\begin{figure}[t]
    \centering
    \includegraphics[width=\linewidth]{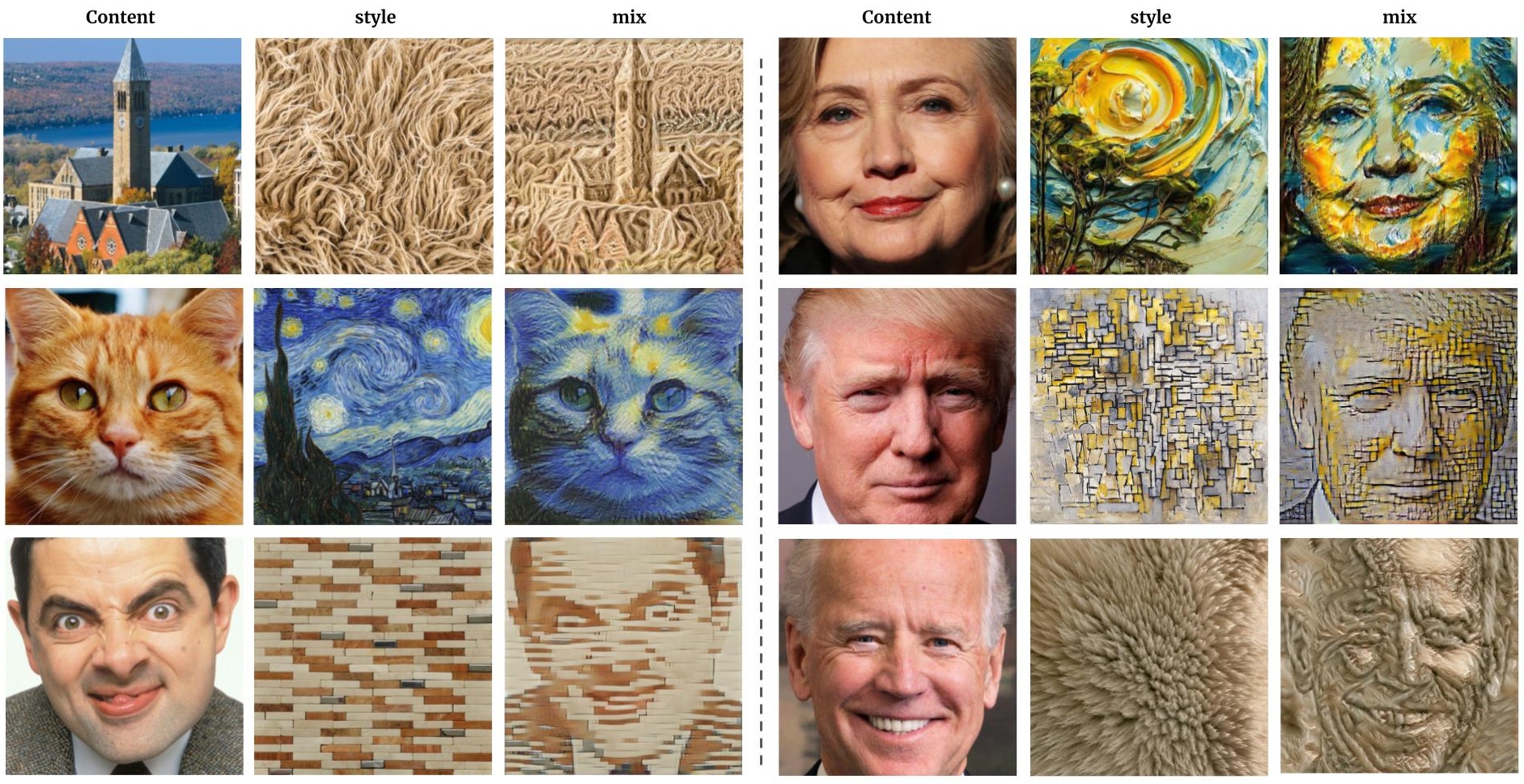}
    \caption{Style/texture transfer: Each image triplet shows a content image, a style/texture image and the results of combining them with our synthesis method.}
\label{fig:Style_Transfer}
\end{figure}

\textbf{Style-transfer}: Figure~\ref{fig:Style_Transfer} shows our results in style transfer: triplets of content, style and the mixed output generated by our method. These results are surprising due to the simplicity of our method. In this task we start the optimization from a rather big image or even use a single-scale configuration. We start the optimization from the content image and match the patch distribution to the style image.

%\subsection{Other}
\textbf{Additional tasks}:
We also applied our method to generating texture images from texture samples. This is done similarly to retargeting but the initial condition is a noise image. Image editing can be performed by first crudely editing the image and then using our algorithm to harmonize its fine details so that it looks real. Figure~\ref{fig:Teaser} shows examples of texture synthesis and image editing is discussed in section \ref{sec:Limitaions}.

Results for additional tasks and more results of reshuffling, retargeting and style-transfer are available in the supplementary material.

\subsection{Efficiency}
\begin{figure}[h]
    \centering
    \includegraphics[width=1\linewidth]{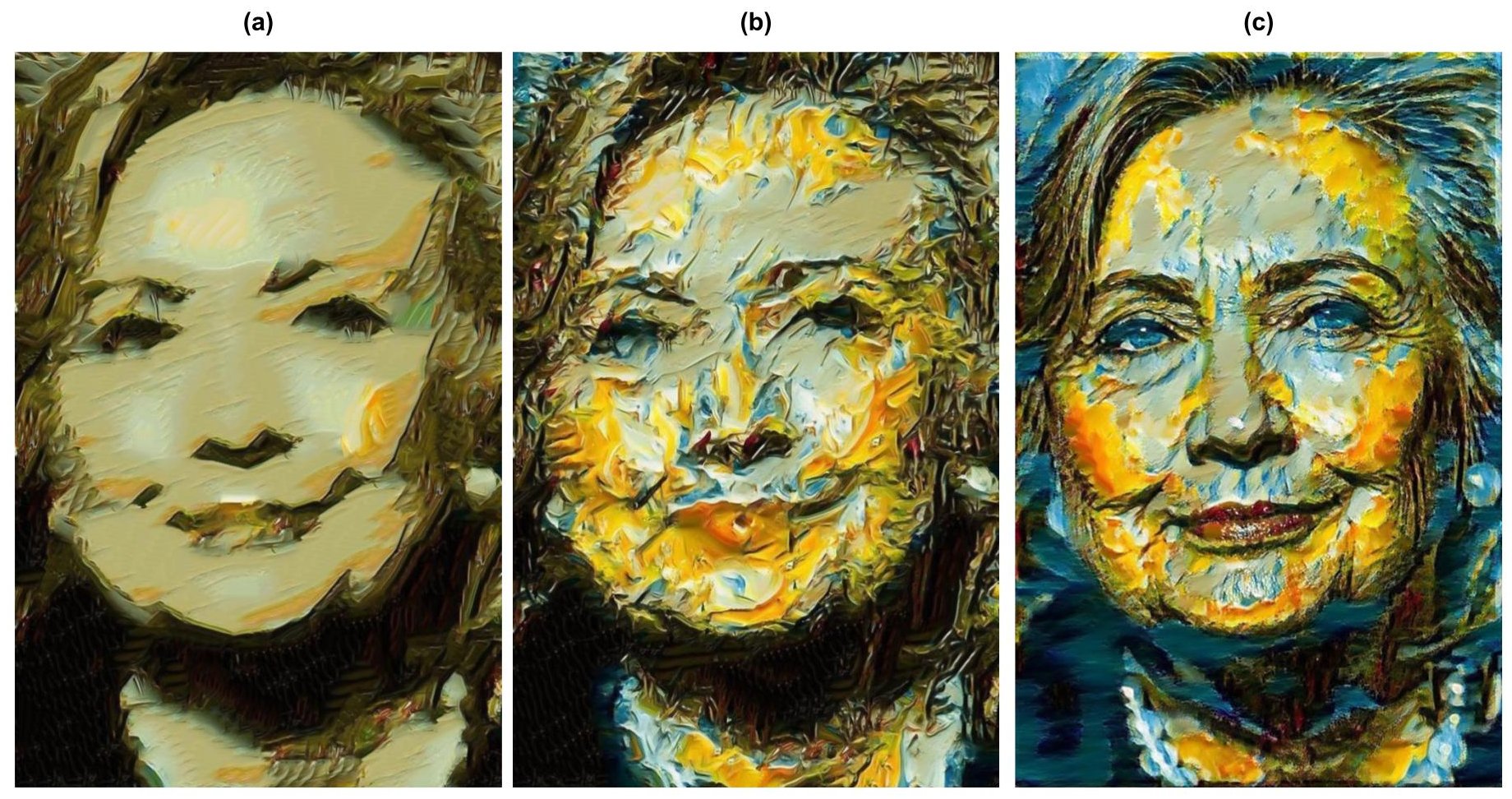}
    \caption{High resolution style transfer images generated (a) GPNN(no-$\alpha$) with approximate nearest neighbor  ($\sim70$ seconds, GPU), (b) GPNN($\alpha$=0.005) with exact nearest neighbor ($\sim1900$ seconds, GPU). (c) Our result ($\sim60$ seconds, GPU). $\alpha$ is the completeness enforcing parameter and no-$\alpha$ means no constraint.}
    \label{fig:HQ_images}
\end{figure}

 A major disadvantage of GAN based methods is the need to train a generator and discriminator for every image. As reported in~\cite{GPNN} this means that generating a new image of size $180 \times 250$ using SinGAN will take about one hour while GPNN (and other methods based on bidirectional similarity) take about two seconds. Our method also does not require any new training and the run times are similar to those of GPNN when the images are small. However, as the size of the image increases, the fact that bidirectional similarity requires computing $M^2$ distances (where $M$ is the number of patches in each image) means that GPNNs run time grows approximately quadratically and generating a novel image of size $1024 \times 678$ with GPNN takes more than half an hour on a GPU. In contrast, our method has complexity of $O(M \log M)$ and takes less than a minute to generate an image of the same size on the same GPU (see figure~\ref{fig:HQ_images}).

The runtime of methods based on bidirectional similarity can be improved by using approximate nearest neighbor search, rather than exact search.
% #TO: Yair: Should we delete this paragraph?
Table~\ref{tab:timing_table} shows an analysis of SWD compute time compared to an exact nearest neighbor search and its approximated counterpart with inverted index~\cite{FAISS}. As can be seen, the use of an inverted index speeds up the search substantially.
%and given that GPDM and GPNN perform 300 and 10 iterations at each image scale, the running time of the two methods are comparable when an approximate search is used. 

\begin{figure}
   \centering
    \begin{tabular}{ |c | c | c | c | c | c |}
    \hline
    image-size & $64^2$ & $128^2$ & $256^2$ & $512^2$ & $1024^2$ \\
    \hline
    SWD(64) & 0.002 & 0.006 & 0.021 & 0.086 & 0.335\\
    NN-exact & 0.086 & 0.110 & 0.336 & 4.144 & 73.23\\
    NN-IVF& 0.101 & 0.137 & 0.382 & 1.082 & 5.530\\
    \hline
\end{tabular}
\captionof{table}{Compute time (seconds) of SWD (64 projections), exact nearest neighbor (FaissFlat) and approximated nearest neighbor (Inverted Index with $\sqrt{M}$ bins where M is the number of patches) for different image sizes. All computation are done on a NVIDIA TITAN X GPU.}
\label{tab:timing_table}

\end{figure}

While in some of our experiments we found that the approximation performs reasonably well, a major disadvantage of using approximate nearest neighbor in the context of GPNN is that it is impractical to modify the similarity metric during the image generation process. Recall that GPNN uses a parameter $\alpha$ to change the similarity of patches based on which patches have already been used but when an inverted index is used, this would require recomputing the inverted index at each iteration and the approximate search would become slower than exact search. 

Figure~\ref{fig:HQ_images} demonstrates this with an example of high resolution style transfer (see figure~\ref{fig:Style_Transfer} for the two input images): while using approximate nearest neighbor (left) speeds up the computation, it requires using $\alpha \rightarrow \infty$ in GPDM and therefore creates lower quality images in which the same patches are reused many times. In contrast,  GPNN with exact nearest neighbor (middle) can use a small $\alpha$ parameter and avoid the reuse of patches. Our method (right) is very fast and since it explicitly optimizes the distance between patch distributions, does a better job of capturing the style  It should be noted that using GPNN with exact nearest neighbor search and with $\alpha \rightarrow \infty$ produces practically the same results as those generated with an approximate search, indicating that the drop in quality is due to the $\alpha$ parameter and not to the inaccuracy of the approximation. We refer the reader to our supplementary material  where one can find further details about GPNN's $\alpha$ parameter and a quantitative evaluation of the accuracy of nearest neighbor approximation methods. 

\textbf{Quality-efficiency tradeoff.} GPDM is an iterative algorithm and there is a natural trade-off between the number of iterations and the quality of the generated images. More random projections at each optimization step and more optimization steps help the patch-distributions to match more closely, ensuring more realistic outputs. 
We refer the interested reader to our supplementary material where we visualize this tradeoff by comparing SIFID (\cite{SINGAN}) scores and average running times on the SIGD16 dataset compared for different number of SWD random projections.

%%%%%%%%% Related Work

\section{Related work} \label{sec:Related Work}
The idea of synthesizing realistic images using patches from a target image goes back to Efros and Leung~\cite{EfrosLeung}. Efros and Freeman~\cite{QUILTING} and Hertzmann et al~\cite{ImageAnalogies} extended this idea to style transfer and other tasks. These classical, nonparametric approaches required finding patch nearest neighbors in high dimensions and a great deal of subsequent work attempted to make the search for nearest neighbors more efficient. The PatchMatch algorithm~\cite{PATCH_MATCH} pointed out that the coherence of patches in natural images can be used to greatly speed up the search for nearest neighbors and this enabled the use of these nonparametric techniques in real-time image editing. Our work is very much inspired by these classical papers but we use the SWD to directly optimize the similarity of patch distributions without computing nearest neighbors. Furthermore, in our approach patches in the synthesized image are not constrained to be direct copies of patches in the training image.

As mentioned in the introduction, a key insight behind successful patch-based methods is the use of some form of bidirectional similarity~\cite{BIDIRECTIONAL}: it is not enough to require that patches in the generated image be similar to patches in the training image. While~\cite{BIDIRECTIONAL} optimized the BDS directly, the GPNN approach~\cite{GPNN} rewards bidirectional similarity indirectly by modifying the similarity measure to penalize patches that have already been used. In a parallel line of work~\cite{PATCH_HIST_EQ_1,PATCH_HIST_EQ_2,PATCH_HIST_EQ_3,PATCH_HIST_EQ_4} attempts are made to make a uniform use of all patches in the source image through patch histogram matching. Our approach optimizes a well-understood similarity (the Sliced Wasserstein Distance) that is guaranteed to be zero only if the two distributions are equal and importantly it optimizes this similarity very efficiently.

SINGAN and InGAN~\cite{SINGAN,INGAN} are both variants of GANs that are trained on a single image with a patch based discriminator. Thus they can be seen as approximately matching the distribution of patches in the generated image and the target image. Our approach directly optimizes the similarity between patch distributions in the two images, requires no training in the traditional sense, and provides superior quality results. More recently \cite{ConSinGAN}, \cite{OneShotGAN} were able to push the performance of SinGAN to match that of our method and that of GPNN but the train time are still orders of magnitude slower than our method. 
%In addition \cite{OneShotGAN} no longer uses a patch discriminator which make it less theoretically interesting for our analysis as discussed in the introduction.

The Sliced Wasserstein Distance was used in a number of image generation tasks but it is most often used to estimate the distance between distributions of full images~\cite{PGGAN}. Thus~\cite{SWD_GAN} train a generative model by replacing the GAN objective with an SWD objective. In contrast, here our focus is on estimating the distance between two patch distributions and we have shown that an unbiased estimate of the distance between patch distributions in the two images can be estimated using a single convolution. 

Although not immediately apparent, neural style transfer \cite{NEURAL_STYLE_IS_MMD} can be seen as a single image generative model that preserves patch distribution. \cite{NEURAL_STYLE} showed that the Gram loss in neural style transfer is equivalent to MMD \cite{MMD} over features of intermediate layers of VGG hence the style transfer objective is to minimize distribution of patches (in size of the layers' receptive fields) between the optimized image and the style target. \cite{SWD_STYLE} also recognized this nature of the Gram loss and suggested replacing it with SWD between the VGG features. Similarly \cite{WCT,WCT_USE_1,WCT_USE_2} use the closed form of the Wasserstein distance between Gaussians to push the spatial distribution in VGG feature maps of a content image into that of a style image and then decode it as a mix image.

The works of \cite{CNNMRF,CONTEXTUAL} closely relate to the neural patch distribution for style transfer. They both suggest objectives that focus on the coherence of neural patches in the synthesized image. Similar to~\cite{GPNN}, they enforce NN similarity of neural-patches but they compute similarity in a pre-trained neural network representations rather than in pixel space. Similar to the classic algorithms, both of these work require explicit computation of patch nearest neighbors. In contrast, our use of SWD allows avoiding the computation of patch nearest neighbors which make our algorithm much faster.

%Another work from which we draw inspiration is \cite{AMIR}. The authors show a connection between neural perceptual losses~\cite{JohnsonAL16} and distance between patch distributions of images and showed that comparing image patch distributions with MMD is a superior image loss to regular pixel-level losses in various vision tasks. We experimented as well with MMD for enforcing image patch distribution but found SWD to work better.

The work of~\cite{SWD_1.5} also use SWD for texture synthesis. They minimize SWD on wavelet coefficients with SGD to synthesize new samples from a given texture image. Our paper differs from theirs in that we work in multiple scales, on pixel-level and compute SWD in a more efficient way. We are able thus to produce much better results and to apply our method to more complicated generative tasks.

%%%%%%%%% Limitations
\section{Limitations and Extensions} \label{sec:Limitaions}
\textbf{Limitations:}
As mentioned previously, approaches based on bidirectional similarity do not explicitly optimize the similarity of patch distributions while our method does. In some applications, the fact that our method attempts to reproduce the relative frequencies of different patches is a limitation. It is often easier to find an image that is coherent (i.e. all patches in the synthesized image come from the target image) than to find one that preserves the relative frequencies. Furthermore, in some tasks such as image editing, the desired image should have a different distribution over patches (e.g. if the user wishes to add a new object to the image). Finally, since our method is based on optimization,  it is not guaranteed to find a good local minima and so the final results sometimes show artifacts as can be seen in figure~\ref{fig:Reshuffling}.

%This means that they may end up synthesizing highly coherent images without being forced to preserve the patch distribution allowing the synthesis of images where only part of the objects from the target image appear or change their relative frequency.

%In contrast, our method attempts to reproduce the relative frequency of different patches and therefore it is unlikely that our generated images will use only part of the target image patches. For example given a half sky half grass target image GPDM will always produce samples with relatively the same number of green and blue patches but BDS based methods can also generate full grass/sky images which one my consider as desirable outputs. For the same reason our method struggles in image editing where the task is to explicitly diverge from the target patch distribution.

\textbf{Extensions:} 
We propose here two extensions to our method. Both extensions are designed  to soften the patch frequency constraint on the output allowing the generation of more diverse outputs.

\begin{figure}%
    \centering
    \subfloat[]{
        \includegraphics[width=0.46\linewidth]{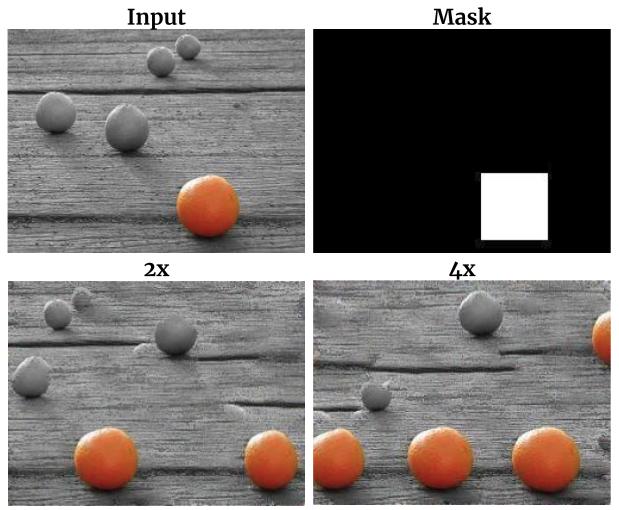}
        \label{fig:Extention_1}
    }%
    \subfloat[]{
        \includegraphics[width=0.46\linewidth]{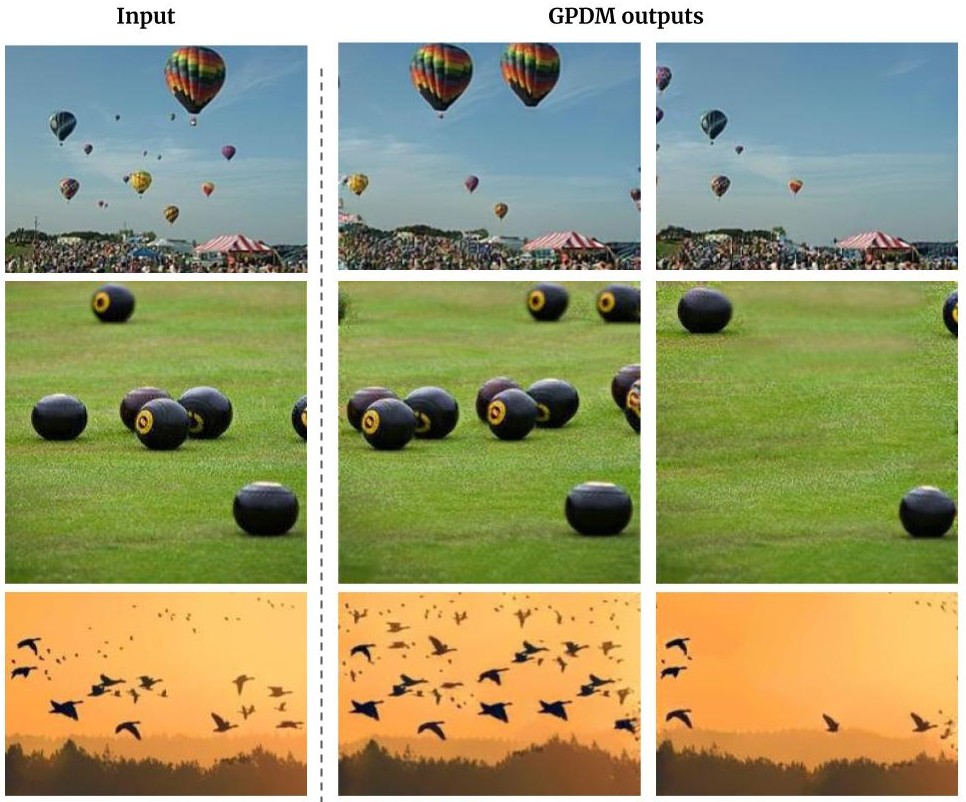}
        \label{fig:Extention_2}
    }%
    \caption{Two Extensions of our method. (a): A mask that specifies target patches whose frequency should be increased is given as an additional input. This allows the user to manipulate the number of objects in the generated images. (b): Batched GPDM. Each row shows the two most diverse outputs out of 10 image batch.}%
    \label{fig:example}%
\end{figure}

The first extension of our method allows the user to explicitly manipulate the distribution over patches that is matched by our algorithm. Figure~\ref{fig:Extention_1} illustrates this. Here the user first manipulates the target distribution of patches by indicating that certain patches should be increased in frequency and the algorithm then generates an image to match the augmented target distribution. As can be seen, this simple modification allows us to generate images with different number of objects based on the user's preference.

The second extension is to use an ensemble of images so that the patch distribution in the ensemble matches that of the target image.  Formally, instead of matching the patch distribution of a single output image to a single target we generate $N$ output images and match their total patch distribution to that of all the patches in $N$ copies of the target image. This way, while the total patch distribution is still strictly preserved, the distribution within each generated image can diverge from that of the original target image. For example when reshuffling 2 images to match the distribution of 2 copies of an image with black balls on grass, one output may contain no balls at all and the other will have twice as many balls while still preserving the frequency of  patches in the two images together. Figure~\ref{fig:Extention_2} shows the results from this algorithm which we call  ``batched-GPD'': For each input image we generated 10 images in parallel and showed the two most different results.

This extension works surprisingly well and yields cleaner and more diverse results. Moreover since the memory footprint of GPDM is linear in the number of patches (in-place sorting) we were able to run in large batches reducing the generation time for a single image.

%% Speculations / Future plans
%In future work we plan to further investigate the potential of batched inference. In particular we will experiment with matching the patch distribution of N \textbf{different} samples from the same scene or even totally different images in an attempt to form a more general generative model.

%%%%%%%%% Conclusions
\section{Conclusion}

Classical approaches for generating realistic images are based on the insight that if we can match the distribution of patches between the generated image and the target image then we will have a realistic image. In this paper we have used the same insight but with a novel twist. We use the Sliced Wasserstein Distance which was recently used as a method to compare distributions of full images when training generative models and we showed that when applied to distributions of patches, an unbiased estimate of the SWD can be computed with a single convolution of the two images. Our experiments shows that this unbiased estimate is sufficient for excellent performance in a wide range of image generation tasks.

\section*{Acknowledgements}
Support from the Israeli Ministry of Science and Technology and the Gatsby Foundation is gratefully acknowledged. We also thank the authors of~\cite{GPNN} for answering our question about their method.

% ---- Bibliography ----
%
% BibTeX users should specify bibliography style 'splncs04'.
% References will then be sorted and formatted in the correct style.
%
%\bibliographystyle{splncs04}
\bibliographystyle{unsrt}
\bibliography{egbib}
\end{document}